\documentclass[10pt,twocolumn,letterpaper]{article}

\usepackage{wacv}              % To produce the CAMERA-READY version
\usepackage{graphicx}
\usepackage{amsmath}
\usepackage{amssymb}
\usepackage{booktabs}
\usepackage{graphicx}
\usepackage{booktabs}
\usepackage{colortbl}
\usepackage[accsupp]{axessibility} 
\usepackage[pagebackref,breaklinks,colorlinks]{hyperref}
\usepackage{orcidlink}
\usepackage{multirow}

\usepackage[pagebackref,breaklinks,colorlinks]{hyperref}

\usepackage{enumitem}

\usepackage[capitalize]{cleveref}
\crefname{section}{Sec.}{Secs.}
\Crefname{section}{Section}{Sections}
\Crefname{table}{Table}{Tables}
\crefname{table}{Tab.}{Tabs.}

\def\wacvPaperID{} % *** Enter the WACV Paper ID here
\def\confName{WACV}
\def\confYear{2025}

\begin{document}

\title{FT2TF: First-Person Statement Text-To-Talking Face Generation}

\author{
Xingjian Diao, Ming Cheng, Wayner Barrios, SouYoung Jin \\
Dartmouth College \\
{\tt\small \{xingjian.diao.gr, ming.cheng.gr, wayner.j.barrios.quiroga.gr, souyoung.jin\}@dartmouth.edu}
}
\maketitle

%%%%%%%%% ABSTRACT
\begin{abstract}
 
Talking face generation has gained immense popularity in the computer vision community, with various applications including AR, VR, teleconferencing, digital assistants, and avatars.
Traditional methods are mainly audio-driven, which have to deal with the inevitable resource-intensive nature of audio storage and processing. 
To address such a challenge, we propose \textbf{FT2TF} -- \underline{\textbf{F}}irst-Person Statement \underline{\textbf{T}}ext-\underline{\textbf{To}}-\underline{\textbf{T}}alking \underline{\textbf{F}}ace Generation, a novel one-stage end-to-end pipeline for talking face generation driven by first-person statement text. 
Different from previous work, our model only leverages visual and textual information without any other sources (e.g., audio/landmark/pose) during inference. 
Extensive experiments are conducted on LRS2 and LRS3 datasets, and results on multi-dimensional evaluation metrics are reported. Both quantitative and qualitative results showcase that FT2TF outperforms existing relevant methods and reaches the state-of-the-art. This achievement highlights our model's capability to bridge first-person statements and dynamic face generation, providing insightful guidance for future work. 
\end{abstract}

\newcommand\blfootnote[1]{%
  \begingroup
  \renewcommand\thefootnote{}\footnote{#1}%
  \addtocounter{footnote}{-1}%
  \endgroup
}

\blfootnote{2025 IEEE/CVF Winter Conference on Applications of Computer Vision (WACV)}

%%%%%%%%% BODY TEXT
\section{Introduction}
\label{sec:intro}

Talking face generation, with the goal of creating a sequence of faces to simulate the appearance of a person talking, has surged in significance amid the metaverse boom and technological advancements. 
This pursuit has gained substantial popularity and importance within the computer vision community, influencing applications including augmented reality~\cite{javornik2022lies}, virtual reality~\cite{fysh2022avatars}, computer games~\cite{shi2020fast, shi2019face}, teleconferencing~\cite{correia2020evaluating, archibald2019using}, video dubbing~\cite{garrido2015vdub, kim2019neural}, special effects for movies~\cite{zhang2022video}, and the development of digital assistants~\cite{thies2020neural} and avatars~\cite{fysh2022avatars, zhou2020makelttalk}.

\begin{figure}[tb]
  \centering
  \includegraphics[width=80mm]{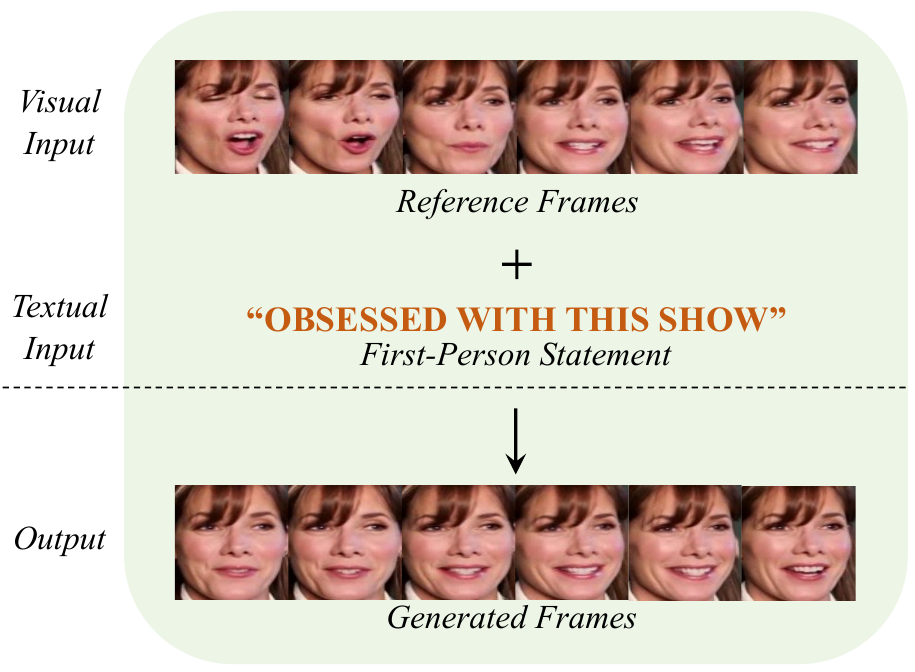}
  \caption{\textbf{
  High-quality text-driven talking face generation.} FT2TF aims to generate realistic talking faces using two inputs: (i) reference talking face frames and (ii) first-person statement text.}
  \label{fig:teaser}
\end{figure}

Traditional methods for talking face generation mainly rely on audio input, which leverages visual and audio cues. These methods, exemplified by works like~\cite{prajwal2020lip,zhou2020makelttalk,zhou2021pose,ji2022eamm,zhong2023identity}, show impressive performance. However, they come with a significant drawback: the need for high-quality audio. Compared to audio-driven methods, text-driven approaches offer unparalleled advantages, notably in terms of:
\begin{itemize}[leftmargin=*]
\item \textbf{Data Storage}: Text data storage requires far less space than audio data. For instance, storing the text of a 300-page article requires a mere 128 KB, while storing audio for the same content demands several gigabytes.
\item \textbf{Data Transmission}: Current audio-driven talking face generation methods necessitate high-quality audio, imposing stringent requirements on data transmission conditions and recording environments \cite{toshpulatov2023talking}. In contrast, text-driven talking face generation methods operate effectively with basic and reliable network transmission conditions.
\end{itemize}

Addressing such challenges holds significant value not only for the scientific community but also for the industry, aiming to mitigate the environmental impact of AI technologies ~\cite{carbonAI}. Consequently, one crucial question has arisen: \textit{Is it feasible to substitute audio with text inputs while ensuring detailed facial expressions of generated frames?}

There exist two primary types of text-driven face generation methods.
The first involves using \textit{third-person} text descriptions where a facial expression is described from an external perspective. These descriptions are leveraged to generate realistic faces using end-to-end GAN architectures \cite{han2022show, ma2023talkclip, yu2023celebv}. 
 The second path employs \textit{first-person} statements which are the speaker’s spoken captions captured in text form. 
While this approach aligns more closely with talking face generation, it has not been thoroughly explored.
Previous methods mainly conduct a two-stage framework including text-to-speech and speech-to-face procedures \cite{oneațua2022flexlip, zhang2022text2video}. These approaches generate audio waveforms from text statements and further generate faces based on synthesized audio \cite{song2022talking}. Therefore,
methods that follow a one-stage end-to-end pattern to generate talking faces driven by first-person statements remain unestablished. Based on this, our objective is to revolutionize the synthesis of coherent and natural talking face videos through first-person statements, encompassing well-synchronized lip movements, accurate cheek actions, detailed textures, and overall facial expressions, as illustrated in Figure~\ref{fig:teaser}. To overcome the challenges above, we propose \textbf{F}irst-Person Statement \textbf{T}ext-\textbf{To}-\textbf{T}alking \textbf{F}ace Generation (\textbf{FT2TF}) -- a one-stage end-to-end pipeline that effectively generates high-quality talking faces from visual-textual cross-modality. Different from previous work, our pipeline encapsulates the entire face generation process without any other sources (\textit{e.g.,} audio, landmarks, poses) being used during inference.

In summary, our contribution is threefold:
\begin{itemize}[leftmargin=*]
    \item {
    \textbf{One-stage End-to-end Continuous-Frame Talking Face Generation.} We propose FT2TF, a one-stage end-to-end pipeline that generates realistic continuous-frame talking faces by integrating visual and textual input.
    }
    \item {
    \textbf{Effective and Efficient Text-Driven Talking Face Generation.} FT2TF utilizes only visual and textual information during inference. It enhances face naturalness, synchronizes lip movements, and improves detailed facial expressions while using fewer trainable parameters.
    }
    \item {
    \textbf{State-of-the-Art Performance.} 
    We conduct extensive experiments on the LRS2 \cite{Afouras18lrs2} and 
LRS3 \cite{afouras2018lrs3} datasets with both qualitative and quantitative results to demonstrate the effectiveness of 
FT2TF. This strongly proves that our model can consistently generate high-quality talking faces across different conditions. 
    }
\end{itemize}

\section{Related Work}
\label{relatedwork}

\subsection{Audio-Driven Talking Face Generation}
The majority of the work on talking face generation is audio-driven which needs both audio and video ground truth as the input, focusing on lip region/whole face generation. 

\noindent
\textbf{Lip Generation.} 
As a groundwork, Chen \textit{et al.} proposes ATVGnet \cite{chen2019hierarchical} that uses audio signal and reference image to generate the region around the lip, while the remaining parts of the face remain static from frame to frame. Based on that, Wav2Lip \cite{prajwal2020lip} generates dynamically changed faces through lip masking and reconstruction strategy. Following the previous lip masking strategy, IP\_LAP \cite{zhong2023identity} works as state-of-the-art by using a Transformer-based \cite{vaswani2017attention} encoder to predict the landmarks for further lip generation. Although these methods generate high-quality dynamic faces, all of them \textbf{\textit{only}} focus on lip generation, failing to capture facial expression details of the entire face.

\noindent
\textbf{Whole Face Generation.}
 Zhou \textit{et al.} \cite{zhou2020makelttalk}
% proposes MakeItTalk \cite{zhou2020makelttalk} which 
disentangles audio sources into content and speaker embeddings, followed by an LSTM-based module for landmark prediction to generate the talking head.
% The predicted landmark is fused with the visual source through an image-to-image translation
% network for the talking-head generation. 
Based on that, models that use input sources other than audio have been proposed. 
PC-AVS \cite{zhou2021pose} utilizes the pose source to control the head pose of the synthesized faces, 
 while EAMM \cite{ji2022eamm} uses pose sequence, emotion source, and audio ground truth to generate the target video manipulated by different emotion guidances. In contrast to existing methods that involve multiple input resources for face generation, our model \textbf{\textit{only}} uses text and reference face frames as the input source. This ensures the efficient generation of high-quality facial representations, free from input requirements such as constraints of audio quality. \looseness-1

\begin{figure*}[htbp]
  \centering
  \includegraphics[width=\textwidth]{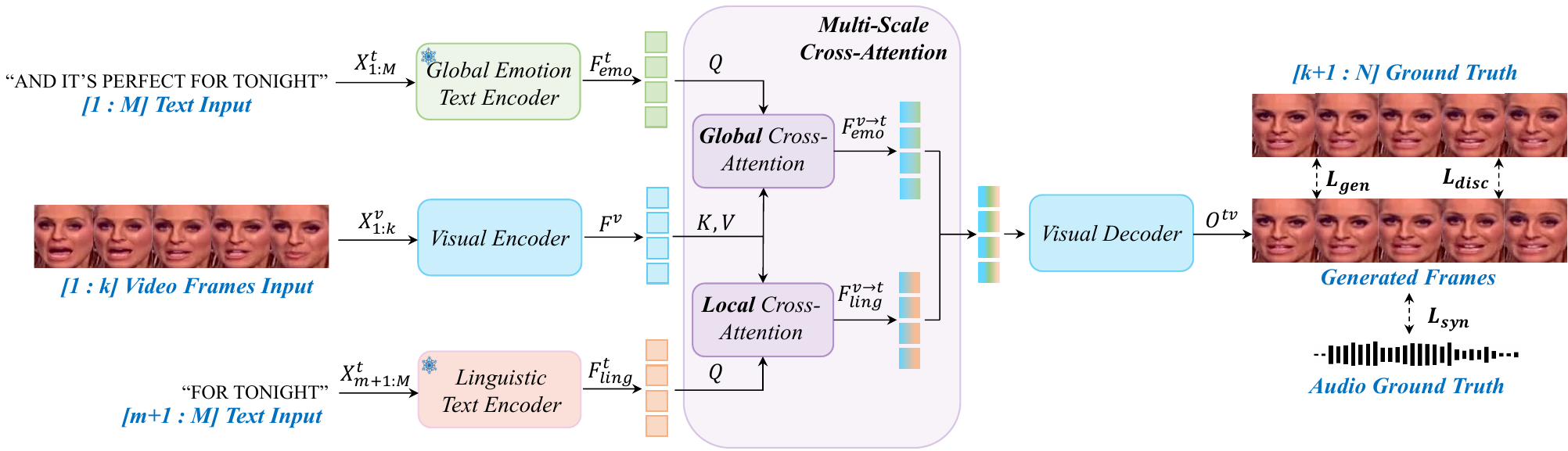}
  \caption{
  \textbf{Overview of FT2TF Pipeline.} The FT2TF pipeline employs two specialized Text Encoders, the Global Emotion Text Encoder, and the Linguistic Text Encoder, for extracting emotional and linguistic text features, respectively. Additionally, a Visual Encoder is utilized to extract visual features. Afterward, it leverages a Multi-Scale Cross-Attention Module for visual-textual fusion. The resulting visual-textual representations are fed to a Visual Decoder to synthesize talking face frames.
  }
  \label{fig:t2f_architecture}
\end{figure*}

\subsection{Text-Driven Talking Face Generation}
The exploration of talking face generation based on first-person statements has not been well-established. Initially,  models using Variational Autoencoder (VAE) \cite{kingma2013auto, li2018video, walker2016uncertain, xue2018visual} and CRNN \cite{balaji2019conditional, tulyakov2018mocogan} are designed to generate video based on text input.
However, the generated frames by these models are unrealistic and over-smoothing, especially in long-sequence video generation tasks \cite{xue2018visual, mo2021long}. Text-driven talking face generation has been divided mainly into two directions: using first-person statements or third-person descriptions. Methods driven by third-person descriptions \cite{han2022show, ma2023talkclip, yu2023celebv} aim to learn the mapping between objective text descriptions and synthesized video. The most relevant approaches using first-person statements are constructed in a two-stage architecture with two separate steps: converting text to speech and then generating faces from the speech \cite{kr2019towards, kumar2017obamanet, yu2019mining, oneațua2022flexlip, zhang2022text2video}. The latest work, TTFS \cite{jang2024faces}, integrates text-to-speech and face-generation modules to synthesize talking face and audio from input text jointly.
Consequently, one-stage methods that generate talking faces directly from first-person text captions remain unexplored \cite{li2021write,liu2022parallel}. Addressing this research gap, our model is constructed as a one-stage end-to-end framework that directly leverages reference talking face frames and first-person text caption to generate high-quality talking faces efficiently. Since relevant exploration has not been widely established, our solution aims to bridge first-person text and dynamic face synthesis to provide insightful guidance for future work. 

\section{FT2TF Pipeline}
\label{sec:method}
We address the challenging problem of generating realistic talking face videos from \textit{first-person} text descriptions. Formally, given $M$ text caption tokens $X^{t}_{1:M}$ as the complete sentence and the initial $k$ talking face frames $X^{v}_{1:k}$ as input, our objective is to synthesize the subsequent talking face frames $X^{v}_{k+1:N}$.  In this context, $t$ represents the textual content, and $v$ represents the visual content. The text captions $X^{t}_{1:M}$ and the talking face frames $X^{v}_{1:N}$ are aligned to correspond to each other.

To tackle the problem, we present a novel pipeline -- FT2TF: First-Person Statement Text-to-Talking Face Generation, as illustrated in Figure~\ref{fig:t2f_architecture}. 
We use two specialized text encoders -- the Global Emotion Text Encoder captures overall emotional tone from the complete text caption $X^{t}_{1:M}$, and the Linguistic Text Encoder encodes semantic compression from the reference text caption $X^{t}_{m+1:M}$ which corresponds to frames $X^{v}_{k+1:N}$. 
Alongside these, a Visual Encoder is employed to extract visual features from reference frames $X^{v}_{1:k}$. Inspired by \cite{li2023blip}, we design and utilize a Multi-Scale Cross-Attention Module that integrates Global and Local Cross-Attention modules to align features from two modalities into the same distribution \cite{vaswani2017attention, li2023blip}, aiming to ensure a comprehensive visual-textual fusion across varying scales and feature dimensions. The resulting visual-textual representations are fed to a Visual Decoder to synthesize talking face frames $X^{v}_{k+1:N}$.

\subsection{Multimodal Encoders}

In this Section, we explore the multimodal encoders utilized in FT2TF. Our investigation concentrates on the design insights and functionalities of the Visual Encoder, Global Emotion Text Encoder, and Linguistic Text Encoder. As shown in Figure~\ref{fig:t2f_architecture}, the features output from these three encoders are denoted as $F^{v}$, $F^{t}_{emo}$, and $F^{t}_{ling}$, respectively. 

\noindent\textbf{Visual Encoder.} 
The CNN-based Visual Encoder plays a pivotal role in extracting image features and facial details from talking face videos. 
Considering the properties of ResBlocks to handle the temporal dynamics and variations in visual input without massive data or strong data augmentations \cite{islam2022recent}, the Visual Encoder is built with ResBlocks as the backbone. The extracted visual features, $F^{v}$, will then be integrated with the textual ones through the Multi-Scale Cross-Attention module. 
Furthermore, Table \ref{tab:backbone_ablation} illustrates the results of various backbones of the Visual Encoder. \looseness=-1 

\noindent\textbf{Global Emotion Text Encoder.} The Global Emotion Text Encoder harnesses the capabilities of the pretrained Emoberta \cite{kim2021emoberta} model. Emoberta serves as the cornerstone for encoding the overarching emotional context embedded within textual descriptions. 
In light of this, it presents itself as a prime contender for global attention. Using the entire face frames, we can direct our exclusive focus toward text emotions and their face representations. 

\noindent\textbf{Linguistic Text Encoder.} We adopt to use the pretrained GPT-Neo~\cite{gao2020pile} as the Linguistic Text Encoder. In our choice to employ this model, we find motivation in the distinctive characteristics of the GPT-Neo, as opposed to the Emoberta~\cite{kim2021emoberta}. GPT-Neo has undergone training using a vast language dataset, which empowers it with enhanced linguistic compression for every statement conveyed by the talking face. Consequently, we intend to narrow our attention solely to the region of the lips, aligned with the phonetic articulation of each word, as can be seen in Figure~\ref{fig:att}.

\subsection{Multi-Scale Cross-Attention and Visual Decoder}
\label{sec_att}
 
Given the encoded textual ($F^{t}_{emo}$, $F^{t}_{ling}$) and visual features ($F^v$), we design a Multi-Scale Cross-Attention module that integrates \textit{Global} and \textit{Local} Cross-Attention modules.  Specifically, features extracted from the Visual Encoder, \(F^{v}\), are used in conjunction with the textual features from the Global Emotion Text Encoder, \(F^t_{emo}\), as input to the Global Cross-Attention. In this context, \(F^{v}\) serves as both the key (\(K\)) and the value (\(V\)), while \(F^t_{emo}\) functions as the query (\(Q\)). A similar process occurs with Local Cross-Attention where \(F^{v}\) serves as $K$ and $V$ and the output of the Linguistic Text Encoder, \(F^t_{ling}\), acts as $Q$.

\begin{figure}[htb]
  \centering
  \begin{subfigure}{\linewidth}
  \centering
    \includegraphics[width=70mm]{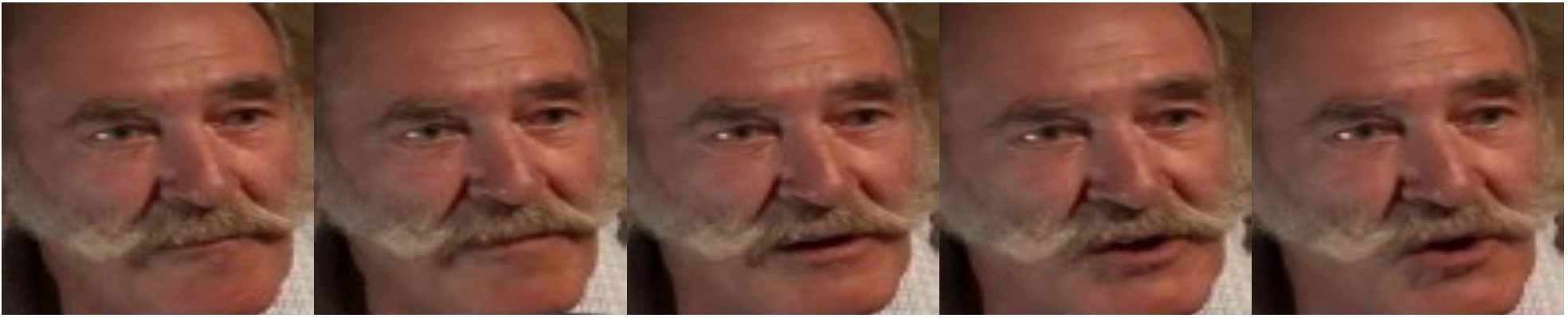}
    \caption{Ground Truth talking face frame sequences.}
    \label{fig:att1}
  \end{subfigure}
  \hfill
  \begin{subfigure}{\linewidth}
  \centering
    \includegraphics[width=70mm]{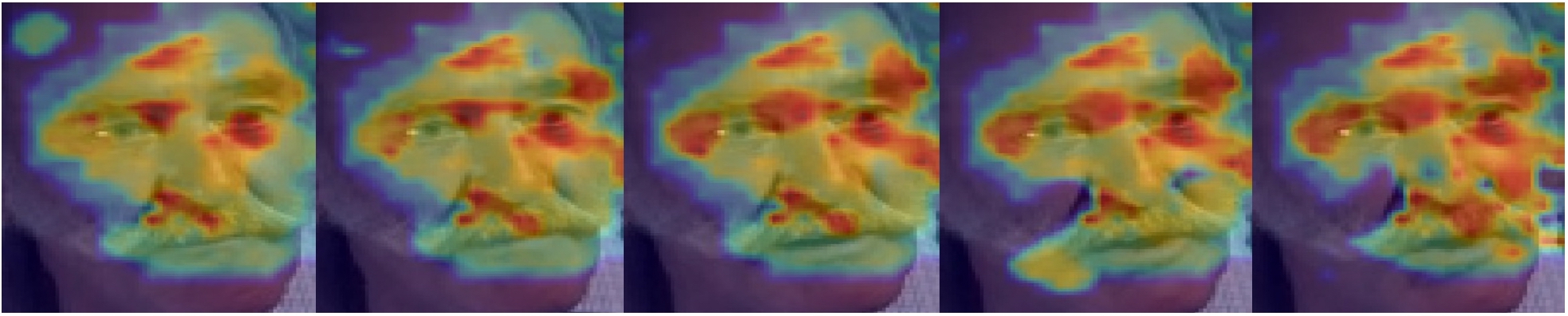}
    \caption{Corresponding attention maps from Global Cross-Attention.}
    \label{fig:att2}
  \end{subfigure}
  \hfill
  \begin{subfigure}{\linewidth}
  \centering
    \includegraphics[width=70mm]{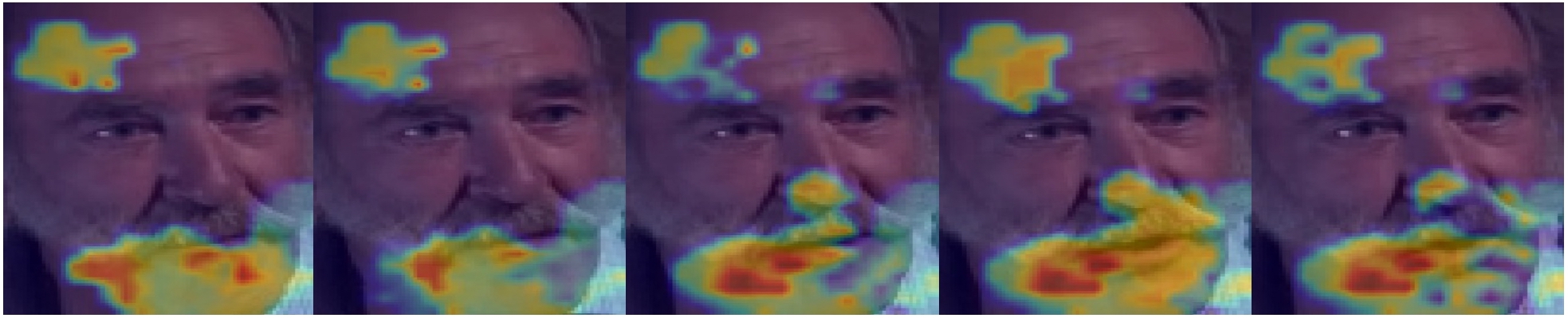}
    \caption{Corresponding attention maps from Local Cross-Attention.}
    \label{fig:att3}
  \end{subfigure}
  \caption{\textbf{Visualization of Global and Local Cross-Attention}. The red areas indicate high attention weights, while the blue areas indicate the opposite.}
  \label{fig:att}
\end{figure}

\noindent\textbf{Global Cross-Attention.} This module serves as a pivotal component for infusing global emotional cues from the textual description into the video context. It combines the encoded video features \(F^v\) with the emotional text description ones \(F^t_{emo}\) and generates the fusion features $F^{v \rightarrow t}_{emo}$. As shown in Figure \ref{fig:att2}, this integration empowers the model to effectively incorporate emotional expression into facial components such as cheeks, eyes, and eyebrows, enabling more expressive and emotionally accurate face synthesis.

\noindent\textbf{Local Cross-Attention.} This module enriches the model's understanding of local language features in the text. Specifically, by fusing the features $F^{v \rightarrow t}_{ling}$ from \(F^v\) and \(F^t_{ling}\), it facilitates the generation of localized and context-aware facial expressions and details. Since the linguistic information is mainly presented by mouth movements, the Local Cross-Attention results in the highest weight in the area around the mouth, as shown in Figure \ref{fig:att3}. This highly enhances the model's ability to capture fine-grained linguistic nuances and improves the overall coherence of the generated face.

\noindent\textbf{Multi-Scale Cross-Attention.}
The collective use of Global and Local Cross-Attention ensures a balanced fusion of global emotional context and local linguistic features. 
As corroborated in Section~\ref{sec:experiment}, this multimodal fusion mechanism enables FT2TF to generate talking face videos with natural lip synchronization and emotionally expressive facial details, resulting in more realistic and engaging outputs.

\noindent\textbf{Alignment.}
In the realm of text-video talking face synthesis, a primary concern revolves around the alignment between textual inputs and video frames. Given the variability in speech rates and temporal intervals between words, especially when considering text-only inputs, we address this challenge by circumventing the alignment problem between text and video frames. Instead, we leverage extensive audiovisual data alongside a highly efficient lip synthesis expert module to ensure the synthesis of text-video talking faces. Operating on an end-to-end basis, our model necessitates correspondence between the video input and the textual content, thereby facilitating the generation of lip images that closely approximate the ground truth during synthesis. As illustrated in Figure \ref{fig:att}, the emotional content and latent linguistic information of the text also influence different facial regions through Multi-Scale Cross-Attention. This phenomenon indicates that our model has successfully aligned features from different distributions.

\noindent\textbf{Visual Decoder.} 
The Visual Decoder is constructed with transpose convolution blocks, aligning each convolutional kernel with its counterpart in the Visual Encoder and matching the channel dimensions to those of \(F^{v\rightarrow t}_{emo}\) and \(F^{v\rightarrow t}_{ling}\). In this way, the output resolution of the Visual Decoder aligns with the model input. The Visual Decoder conducts channel-wise convolution operations on the two text features and visual features, aiming to reconstruct the visual information embedded in the text features.

The output of the Global and Local Cross-Attention (\(F^{v\rightarrow t}_{emo}\), \(F^{v\rightarrow t}_{ling}\)) are concatenated and subsequently input into the Visual Decoder. The  Visual Decoder generates faces from the \(k+1\) frame to the \(N\) frame \(O^{tv}_{k+1}, ..., O^{tv}_{N}\). 

\subsection{Pipeline Training}

Three key loss functions are constructed for pipeline training. 

\noindent \textbf{Generation Loss.} The generation loss, $L_{gen}$, measures the pixel-wise discrepancy between the generated frames and the ground truth ones, with the length represented by $N_{gen}$. This loss primarily focuses on supervising the accurate generation of pixel-level details to ensure the fidelity of the synthesized frames, as shown below:

\begin{equation}
L_{gen} = \frac{1}{N_{gen}} \sum_{i=1}^{N_{gen}} \left\| X^{v}_{k+i} - O^{tv}_{k+i} \right\|_1.
\end{equation}

\noindent \textbf{Synthesis Loss.} To improve lip synchronization and the naturalness of lip movements~\cite{prajwal2020lip}, the synthesis loss, $L_{syn}$, leverages the audio component by integrating the Mel-spectrum generated by the Face Synthesizer. While it is not explicitly expressed in mathematical terms, $L_{syn}$ involves the alignment of the synthesized talking face frames with the corresponding audio data $A_{k+i}$:

\begin{equation}
L_{syn} = \frac{1}{N_{gen}} \sum_{i=1}^{N_{gen}} SyncNet(O^{tv}_{k+i}, A_{k+i}).
\end{equation}

\noindent \textbf{Discriminator Loss.} The discriminator loss \cite{goodfellow2014generative}, $L_{disc}$, aims to guide the encoder-decoder process to produce more realistic and coherent images. It operates by utilizing a discriminator that provides binary predictions $O_{pred}$ based on whether the input is the generated one or the ground truth. Specifically, the corresponding labels $Y_{disc}$ are assigned the label 1 for the generated data and 0 for the ground truth, producing a binary cross-entropy loss:

\begin{equation}
\begin{split}
L_{disc} = -\frac{1}{N_{gen}} \sum_{i=1}^{N_{gen}} (&Y_{disc} \cdot \log(O_{pred})\\
&+ (1 - Y_{disc}) \cdot \log(1 - O_{pred})).
\end{split}
\end{equation}

Eventually, the total loss $L_{total}$ is expressed as follows:

\begin{equation}
L_{total} = \lambda_{1} L_{gen} + \lambda_{2} L_{syn} + \lambda_{3} L_{disc}.
\end{equation}

\section{Experiment}
\label{sec:experiment}

\subsection{Datasets}
\label{intro_dataset}
We conduct experiments on two widely used and substantial large-scale text-audio-visual datasets: LRS2~\cite{Afouras18lrs2} and LRS3~\cite{afouras2018lrs3}.

\begin{itemize}[leftmargin=*]
\item \textbf{LRS2} \cite{Afouras18lrs2}: The dataset comprises 48,164 video clips sourced from BBC outdoor broadcasts, divided into training, validation, and test sets with 45,839, 1,082, and 1,243 clips, respectively. Each clip is paired with a text caption of up to 100 characters. To ensure uniform input lengths, we retain videos with frame counts between 30 and 35.

\item \textbf{LRS3} \cite{afouras2018lrs3}: The dataset comprises over 400 hours of content from 5,594 indoor TED/TEDx talks, totaling 151,819 videos with corresponding text captions. Specifically, it includes 32,000 training/validation utterances and 1,452 test utterances. Consistent with the LRS2 dataset, we retain videos with frame counts between 30 and 35.
\end{itemize}

Although the two datasets share similar names, it is important to emphasize that they have \textbf{\textit{no overlap}}. The LRS2 and LRS3 videos differ significantly in terms of scenes, data collection methods, character actions, environmental conditions, and other attributes.

\subsection{Comparison Methods}

We conduct a comprehensive evaluation of our method against state-of-the-art approaches for talking face generation. To demonstrate the effectiveness and efficiency of our model, we compare it with state-of-the-art audio-driven and text-driven models on the LRS2 and LRS3 datasets. These models are categorized based on two criteria:

\begin{itemize}[leftmargin=*]
\item \textbf{Generation Task}: Methods are categorized into either \textit{Lip Region} or \textit{Whole Face}, with the former generating only the lip region and the latter generating the entire face.  

\item \textbf{Modality}: Methods are classified based on their input modality, including \textit{Audio+Video (\textit{A+V})}, 
\textit{Text+Video (\textit{T+V})}, \textit{Audio+Video+Pose/Landmarks (A+V+P/L)}.
\end{itemize}

\subsection{Evaluation Metrics}
We employ multiple evaluation metrics across various criteria, ranging from evaluating video synthesis quality to evaluating facial expression representation. 
To evaluate the similarity between generated results and ground truth ones, we apply Peak Signal-to-Noise Ratio (PSNR) and Structural Similarity (SSIM) \cite{wang2004image}. 
In addition, we utilize Learned Perceptual Image Patch Similarity (LPIPS) \cite{zhang2018unreasonable} and Frechet Inception Distance (FID) \cite{heusel2017gans} for assessing feature-level similarity.
Considering identity preservation, we leverage cosine similarity (CSIM) between identity vectors extracted by the ArcFace \cite{deng2019arcface} network. 
Moreover, aligning with the approach outlined in \cite{xie2021towards}, we measure the normalized lip landmarks distance (LipLMD) between generated and ground truth frames to evaluate lip synchronization quality. The scores for all methods are calculated over the \textit{\textbf{entire face}} for a fair comparison.

\begin{table*}[]
\centering
\resizebox{0.9\textwidth}{!} {
\begin{tabular}{cccccccccc}
\toprule
\multicolumn{1}{c}{\textbf{Method}}     & \multicolumn{1}{c}{\textbf{Dataset}} & \multicolumn{1}{c}{\textbf{Generation}} & \multicolumn{1}{c}{\textbf{Input}} & \textbf{PSNR}$\uparrow$        & \textbf{SSIM}$\uparrow$        & \textbf{LPIPS}$\downarrow$       & \textbf{FID}$\downarrow$         & \textbf{LipLMD}$\downarrow$      & \textbf{CSIM}$\uparrow$        \\ 
\midrule
\multicolumn{1}{c|}{ATVGnet \cite{chen2019ATVGnet}}              & \multicolumn{1}{c|}{}                 & \multicolumn{1}{c|}{Lip Region}                 & \multicolumn{1}{c|}{A+V}            & 11.55                & 0.3944               & 0.5575               & 223.26               & 0.27401              & 0.1020               \\
\multicolumn{1}{c|}{Wav2Lip \cite{prajwal2020lip}}              & \multicolumn{1}{c|}{}                 & \multicolumn{1}{c|}{Lip Region}                 & \multicolumn{1}{c|}{A+V}            & 27.92                & 0.8962               & 0.0741               & 43.46                & 0.02003              & 0.5925               \\
\multicolumn{1}{c|}{IP\_LAP \cite{zhong2023identity}}              & \multicolumn{1}{c|}{}                 & \multicolumn{1}{c|}{Lip Region}                 & \multicolumn{1}{c|}{A+V+L}            & 32.91                & 0.9399               & 0.0303               & 27.87                & \textbf{{0.01293}}     & 0.6523               \\ 
\cmidrule{1-1}
\cmidrule{3-10} 
\multicolumn{1}{c|}{EAMM \cite{ji2022eamm}}                 & \multicolumn{1}{c|}{LRS2}             & \multicolumn{1}{c|}{Whole Face}                 & \multicolumn{1}{c|}{A+V+P}            & 15.17                & 0.4623               & 0.3398               & 91.95                & 0.15191              & 0.2318               \\
\multicolumn{1}{c|}{PC-AVS \cite{zhou2021PC-AVS}}               & \multicolumn{1}{c|}{}                 & \multicolumn{1}{c|}{Whole Face}                 & \multicolumn{1}{c|}{A+V+P}            & 15.75                & 0.4867               & 0.2802               & 110.60               & 0.07569              & 0.3927               \\
\multicolumn{1}{c|}{MakeItTalk \cite{zhou2020makelttalk}}           & \multicolumn{1}{c|}{}                 & \multicolumn{1}{c|}{Whole Face}                 & \multicolumn{1}{c|}{A+V}            & 17.25                & 0.5562               & 0.2237               & 76.57                & 0.05024              & 0.5799               \\
\multicolumn{1}{c|}{TTFS \cite{jang2024faces}}           & \multicolumn{1}{c|}{}                 & \multicolumn{1}{c|}{Whole Face}                 & \multicolumn{1}{c|}{T+V}            & 30.28                & 0.9436               & 0.0173               & \cellcolor{gray!25}\textbf{19.62}                & {0.03184}              & 0.8149               \\
\multicolumn{1}{c|}{\textbf{FT2TF (Ours)}} & \multicolumn{1}{c|}{}                 & \multicolumn{1}{c|}{Whole Face}                 & \multicolumn{1}{c|}{T+V}            &  \cellcolor{gray!25}\textbf{33.20}       & \cellcolor{gray!25}\textbf{0.9901}      & \cellcolor{gray!25}\textbf{0.0058}      & {22.52}       & \cellcolor{gray!25}{0.02356}              & \cellcolor{gray!25}\textbf{0.9642}      \\ 
\midrule
\multicolumn{1}{c|}{ATVGnet \cite{chen2019ATVGnet}}              & \multicolumn{1}{c|}{}                 & \multicolumn{1}{c|}{Lip Region}                 & \multicolumn{1}{c|}{A+V}            & 10.90                & 0.3791               & 0.5667               & 190.49               & 0.30564              & 0.1176               \\
\multicolumn{1}{c|}{Wav2Lip \cite{prajwal2020lip}}              & \multicolumn{1}{c|}{}                 & \multicolumn{1}{c|}{Lip Region}                 & \multicolumn{1}{c|}{A+V}            & 28.45                & 0.8852               & 0.0683               & 49.60                & 0.02001              & 0.5909               \\
\multicolumn{1}{c|}{IP\_LAP \cite{zhong2023identity}}              & \multicolumn{1}{c|}{}                 & \multicolumn{1}{c|}{Lip Region}                 & \multicolumn{1}{c|}{A+V+L}            &\textbf{32.97}      & 0.9222               & 0.0310               & 29.96                & \textbf{0.01353}     & 0.6385               \\ 
\cmidrule{1-1} \cmidrule{3-10} 
\multicolumn{1}{c|}{EAMM \cite{ji2022eamm}}                 & \multicolumn{1}{c|}{LRS3}             & \multicolumn{1}{c|}{Whole Face}                 & \multicolumn{1}{c|}{A+V+P}            & 15.37                & 0.4679               & 0.3812               & 108.83               & 0.17818              & 0.2689               \\
\multicolumn{1}{c|}{PC-AVS \cite{zhou2021PC-AVS}}               & \multicolumn{1}{c|}{}                 & \multicolumn{1}{c|}{Whole Face}                 & \multicolumn{1}{c|}{A+V+P}            & 15.60                & 0.4732               & 0.3321               & 115.25               & 0.10611              & 0.3537               \\
\multicolumn{1}{c|}{MakeItTalk \cite{zhou2020makelttalk}}           & \multicolumn{1}{c|}{}                 & \multicolumn{1}{c|}{Whole Face}                 & \multicolumn{1}{c|}{A+V}            & 17.78                & 0.5607               & 0.2788               & 97.99                & 0.08432              & 0.5465               \\
\multicolumn{1}{c|}{TTFS \cite{jang2024faces}}           & \multicolumn{1}{c|}{}                 & \multicolumn{1}{c|}{Whole Face}                 & \multicolumn{1}{c|}{T+V}            & 29.75                & 0.9317               & 0.0206               & \cellcolor{gray!25}\textbf{13.98}                & 0.03061              & 0.8243               \\
\multicolumn{1}{c|}{\textbf{FT2TF (Ours)}} & \multicolumn{1}{c|}{}                 & \multicolumn{1}{c|}{Whole Face}                 & \multicolumn{1}{c|}{T+V}            & \cellcolor{gray!25}{31.63}               & \cellcolor{gray!25}\textbf{0.9895}      & \cellcolor{gray!25}\textbf{0.0085}      & {15.97}       & \cellcolor{gray!25}0.02545              & \cellcolor{gray!25}\textbf{0.9665}      \\ 
\bottomrule
\end{tabular}
}
\caption{
\textbf{Comparison with state-of-the-art talking face generation methods on different evaluation metrics on LRS2 and LRS3 Datasets.}
\textbf{Bold} results indicate the best among \textit{\textbf{all}} methods, while the results in the \colorbox{gray!25}{gray} background indicate the best among \textit{\textbf{whole face generation}} methods. ``V'' - Visual Modality. ``A'' -  Audio Modality. ``T'' - Textual Modality. ``L'' - Landmark Source. ``P'' - Pose Source.
}
\label{tab:results}
\end{table*}

\subsection{Comparison with State-of-the-art}

\subsubsection{Quantitative Results}

As shown in Table \ref{tab:results}, FT2TF achieves state-of-the-art performance across multiple evaluation metrics for talking face generation, as evidenced by the quantitative results on both LRS2 and LRS3 datasets.

\noindent\textbf{Video Quality Enhancement.} 
FT2TF achieves state-of-the-art results across video quality evaluation metrics (PSNR, SSIM, and LPIPS). For example, on the LRS2 dataset, FT2TF outperforms TTFS \cite{jang2024faces}, the current state-of-the-art T+V method, with a $9.6\%$ improvement in PSNR, a $4.9\%$ improvement in SSIM, and a significantly lower LPIPS value, reduced by $66.5\%$. Moreover, FT2TF consistently outperforms lip-specific methods, even those using ground truth for the remaining parts of the face, further validating its superior performance. Similar results are observed on the LRS3 dataset, highlighting FT2TF's ability to generate talking faces with high-quality and visual realism, closely resembling real-world data.

\noindent
\textbf{Data Distribution Alignment.}
In terms of FID, which evaluates the alignment of data distributions, FT2TF achieves scores of $22.52$ on LRS2 and $15.97$ on LRS3, demonstrating competitive performance compared to state-of-the-art methods \cite{jang2024faces}. The low FID values indicate that FT2TF generates frames with distributions statistically close to real ones, underscoring the model's capacity to produce natural-looking talking faces.

\begin{figure}[tb]
  \centering
  \includegraphics[width=0.4\textwidth]{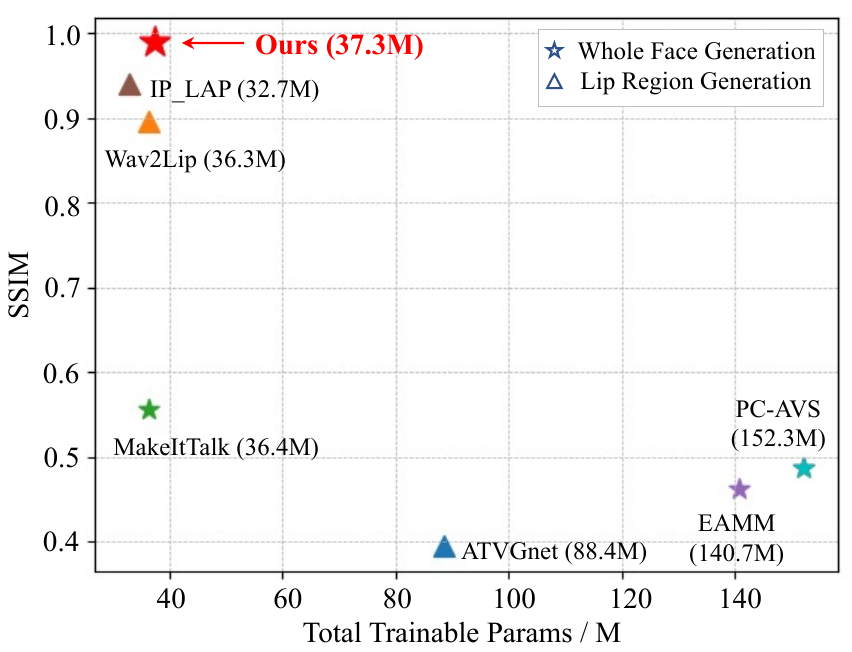}
  \caption{\textbf{Efficiency comparison.} Our analysis includes a comparison of SSIM scores against the total trainable parameters for various models, demonstrating our model's efficiency.}
  \label{fig:ssim/params}
\end{figure}

\noindent
\textbf{Lip Synchronization and Identity Preservation.}
Although FT2TF is not specifically designed for lip generation, it achieves comparable lip synchronization performance (LipLMD) to the state-of-the-art lip-specific method, IP\_LAP \cite{zhong2023identity}, and surpasses all whole-face generation methods. Furthermore, FT2TF excels in identity preservation, as evidenced by the highest CSIM scores. This underscores its capability to maintain accurate facial features and expressions across frames, ensuring the realistic and coherent generation of talking faces.

\noindent
\textbf{Robustness Across Datasets.} FT2TF demonstrates consistent and robust performance across distinct and non-overlapping datasets. Despite the variations between the two datasets described in Section \ref{intro_dataset}, FT2TF maintains stable, high-quality generation capabilities. This underscores the model's adaptability and its strong generalization ability across diverse video sources.

\begin{figure*}[htbp]
  \centering
    \includegraphics[width=140mm]{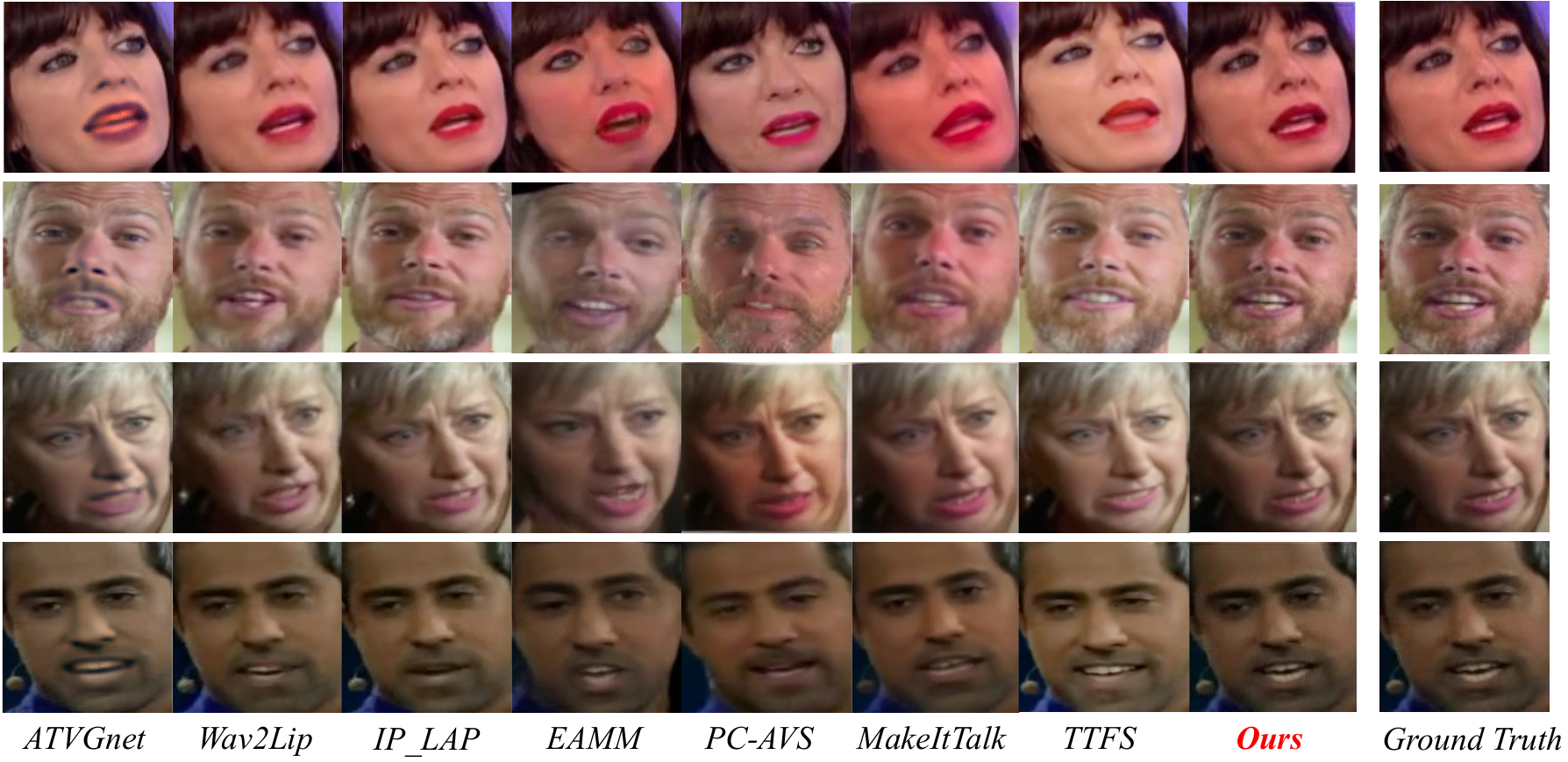}
    \caption{\textbf{Qualitative comparison with state-of-the-art methods on LRS2 and LRS3.} The three models on the left specialize in lip generation, whereas the others are designed to generate entire faces. TTFS \cite{jang2024faces} and our model are text-driven, whereas the remaining methods are audio-driven. Our model consistently generates the most detailed and accurate talking faces across diverse roles, genders, and ages under different lighting and head-poses conditions.}
  \label{fig:visual_single_comp}
\end{figure*}

\begin{figure*}[htbp]
  \centering
    \includegraphics[width=150mm]{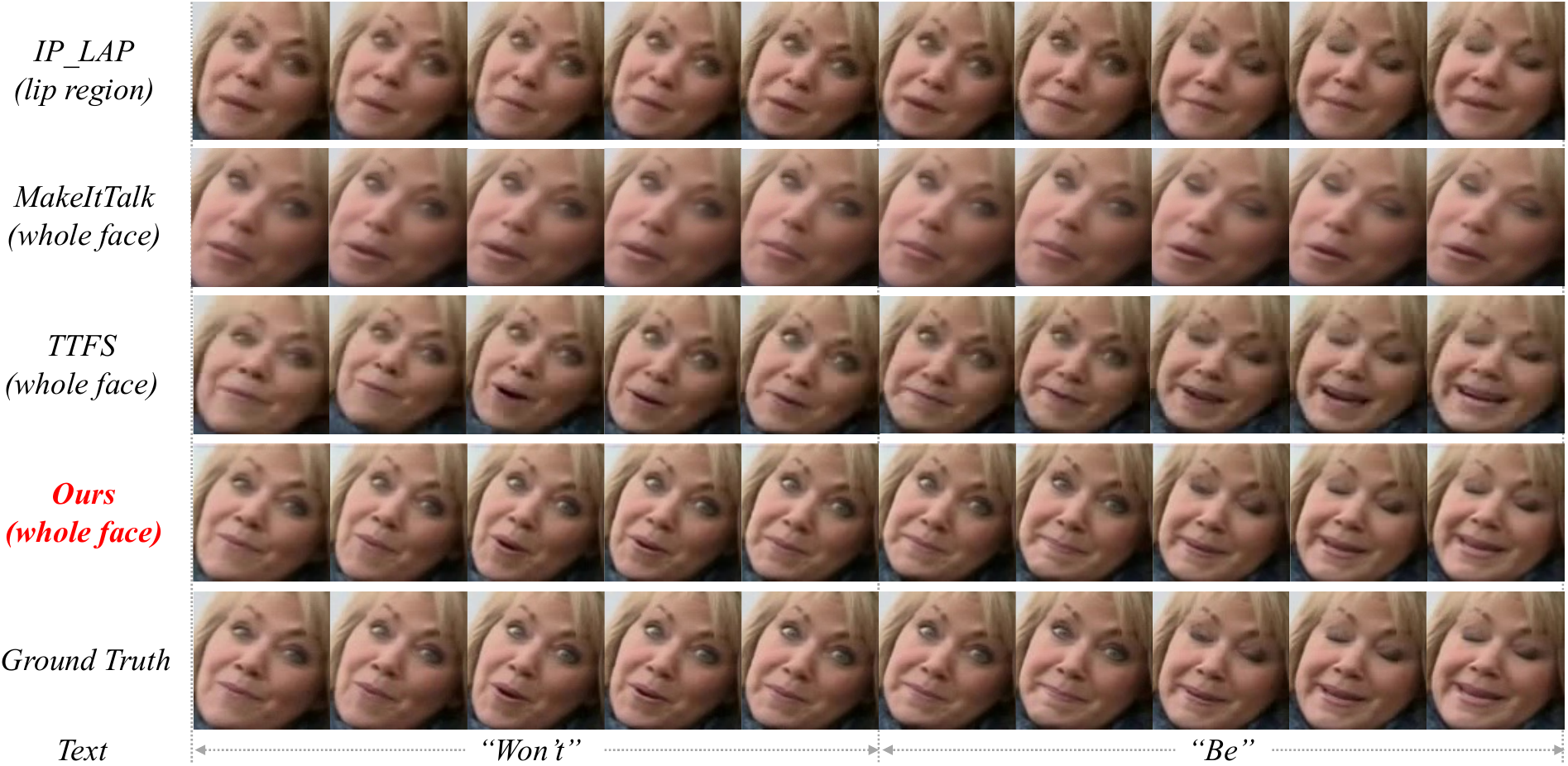}
    \caption{
    \textbf{
    Qualitative comparison demonstration in continuous talking face frames Generation.} Our model demonstrates enhanced capability in generating talking faces with temporally coherent and accurately synchronized lip movements that align with the spoken text.
    }
  \label{fig:long_seq_res}
\end{figure*}

\noindent
\textbf{Efficiency Analysis.} 
Figure \ref{fig:ssim/params} compares the total trainable parameters across various models, including both lip-specific and face-generation approaches. 
Specifically, compared to the
state-of-the-art methods that only learn lip-around features instead of the whole face (IP\_LAP \cite{zhong2023identity}, Wav2Lip \cite{prajwal2020lip}), our model achieves a higher SSIM score while using a competitive number of trainable parameters. 
Moreover, our model outperforms other cutting-edge face generation models significantly. Quantitatively, we achieve a $78.01\%$ improvement in SSIM with only $2.47\%$ more trainable parameters than MakeItTalk \cite{zhou2020makelttalk}. In comparison to other face-generation models listed in the figure, we present substantial advantages in terms of both trainable parameters and performance, with an average performance improvement of $122.88\%$ and an average parameter decrease of $68.94\%$.

\subsubsection{Qualitative Results}
We process all talking face frames using the same face detector \cite{prajwal2020lip}, enabling a more detailed comparison of facial expressions. Figure \ref{fig:visual_single_comp} shows the comparison of generation quality between our model and the relevant methods on LRS2 and LRS3. Our results stand out as the most similar to ground truth in diverse conditions, including different character roles and identities, various lighting, and face angles.
Benefiting from the Global Emotion Text Encoder, Linguistic Text Encoder, and Multi-Scale Cross-Attention modules, our model accurately preserves facial expression details, such as teeth, face shape, head pose, skin color, and texture details.
Furthermore, we evaluate multiple consecutive video frames, as shown in Figure \ref{fig:long_seq_res}. Specifically, a temporal comparison with the state-of-the-art face generation methods shows that our model constructs the most accurate and consistent video frames in the time domain. In addition to the visual demonstration, we also report quantitative results of temporal generation in Section \ref{sec:user_study}.

\subsection{User Study}
\label{sec:user_study}
We conduct a user study to evaluate videos synthesized by state-of-the-art lip and face generation methods. 30 participants with diverse roles, genders, and ages are invited to watch 4 videos from each method and evaluate on a scale of 1 to 5 for four different aspects: temporal transition smoothness, lip synchronization accuracy, facial detail naturalness, and overall video quality. According to the Mean Opinion Scores (MOS) reported in Table \ref{table_userstudy}, our model outperforms the state-of-the-art lip/face generation methods from all evaluation domains.

\begin{table}[t]
\centering
\resizebox{0.9\columnwidth}{!}{%
\begin{tabular}{ccccc}
\toprule
\textbf{Method} & \textbf{\begin{tabular}[c]{@{}c@{}}Temp\\ Change\end{tabular}} & \textbf{\begin{tabular}[c]{@{}c@{}}Lip\\ Sync\end{tabular}} & \textbf{\begin{tabular}[c]{@{}c@{}}Facial\\ Details\end{tabular}} & \textbf{\begin{tabular}[c]{@{}c@{}}Video\\ Quality\end{tabular}} \\ 
\midrule
MakeItTalk \cite{zhou2020makelttalk}       & 1.83                                                      &           1.26                                                   &          1.70                                                          &    1.87                                                              \\
IP\_LAP \cite{zhong2023identity}          & 3.13                                                        &         2.90                                                     &      2.99                                                              &     2.82                                                              \\
% \midrule
\textbf{FT2TF (Ours)}    &      \textbf{4.50}                                                             &       \textbf{4.58}                                                       &      \textbf{4.77}                                                              & \textbf{4.43}                                                                  \\ 
\bottomrule
\end{tabular}}
\caption{
\textbf{User study on the state-of-the-art lip/face generation methods.} Mean Option Scores (MOS) are reported based on four aspects:
\textbf{Temp Change:} temporal transition smoothness.
\textbf{Lip Sync:} lip synchronization accuracy.
\textbf{Facial Details:} facial detail naturalness.
\textbf{Video Quality:} overall video quality.
We conducted this experiment over three different methods.
}
\label{table_userstudy}
\end{table}

\subsection{Ablation Analysis}
\label{sec:ablation}

We conduct a comprehensive analysis of the ablation studies on the FT2TF to investigate the impact of key components, including the Cross-Attention mechanisms, loss terms, and different visual encoder backbones. The primary goal is to analyze the model's performance and demonstrate the contribution of each component in generating high-quality face frames from first-person text statements.

\begin{table}[h]
\centering
\resizebox{0.9\columnwidth}{!}{%
\begin{tabular}{ccccc}
\toprule
\textbf{Method} & \textbf{PSNR}$\uparrow$ & \textbf{SSIM}$\uparrow$ & \textbf{FID}$\downarrow$ & \textbf{CSIM}$\uparrow$\\
\midrule
 w/o Global CA & 23.03 & 0.9101 & 40.59 & 0.8965\\
 w/o Local CA & 21.49 & 0.8975 & 40.56 & 0.8925\\
 w/o Both CA & 22.73 & 0.7317 & 127.38 & 0.5682\\
 {\textbf{FT2TF} \textbf{(Ours)}} & \textbf{33.20} & \textbf{0.9901} & \textbf{22.52} & \textbf{0.9642} \\
 \bottomrule
\end{tabular}
}
\caption{\textbf{Ablation study on Cross-Attention mechanisms.} 
We conduct experiments on the Cross-Attention (CA) mechanisms on LRS2. 
}
\label{tab:total_ablation}
\end{table}

\noindent
\textbf{{Cross-Attention Mechanisms.}}
The ablation study on the Cross-Attention module is shown in Table \ref{tab:total_ablation}. By removing either Global or Local Cross-Attention, a significant drop in all the metrics can be observed. Since emotional coherence and linguistic information are enhanced from the Cross-Attention module as discussed in Section \ref{sec_att}, the model without it fails to generate high-quality and distribution-aligned faces with the person's identity being preserved.

\begin{table}[h]
\centering
\resizebox{0.9\columnwidth}{!}{%
\begin{tabular}{ccccc}
\toprule
\textbf{Method} & \textbf{PSNR}$\uparrow$ & \textbf{SSIM}$\uparrow$ & \textbf{FID}$\downarrow$ & \textbf{CSIM}$\uparrow$\\
\midrule
 w/o $L_{syn}$ & 28.92 & 0.9389 & 40.59 & 0.8965\\
 w/o $L_{disc}$ & 19.90 & 0.8736 & 25.66 & 0.9305\\
\textbf{FT2TF (Ours)} & \textbf{33.20} & \textbf{0.9901} & \textbf{22.52} & \textbf{0.9642}\\
\bottomrule
\end{tabular}
}
\caption{\textbf{Ablation study on loss terms.} Specifically, we conduct ablation study on $L_{syn}$ and $L_{disc}$ on LRS2.}
\label{tab:loss_ablation}
\end{table}
\noindent
\textbf{{Role of Loss Terms.}} We delve into the role of specific loss terms, $L_{syn}$ and $L_{disc}$, as shown in Table \ref{tab:loss_ablation}.
$L_{syn}$ focuses on improving the naturalness of lip movements and facial expression details, which leads to a significant gain in all quantitative metrics. $L_{disc}$ guides the model to generate more realistic frames through binary predictions by the discriminator, contributing to accurate and expressive faces. 

\begin{table}[h]
\centering
\resizebox{0.85\columnwidth}{!}{%
\begin{tabular}{ccccc}
\toprule
 \textbf{Backbone} & \textbf{PSNR}$\uparrow$ & \textbf{SSIM}$\uparrow$ & \textbf{FID}$\downarrow$ & \textbf{CSIM}$\uparrow$\\
\midrule
 ViT-B &  31.30 &  0.9690  & 27.61  & 0.9378\\
 ViT-L &  32.89 &  0.9690  & \textbf{20.60}  & 0.9434\\
 ResBlocks & \textbf{33.20} & \textbf{0.9901} & 22.52 & \textbf{0.9642}\\
\bottomrule
\end{tabular}
}
\caption{\textbf{ Ablation study on Visual Encoder backbones.} We construct the model with ViT/ResBlocks as the encoder backbone on LRS2.
}
\label{tab:backbone_ablation}
\end{table}

\noindent
\textbf{{Encoder Backbones.}} We investigate the effectiveness of different backbone structures of the Visual Encoder, including ViT with different scales (ViT-B/L) and ResBlocks. As shown in Table \ref{tab:backbone_ablation}, our model achieves stable and effective results with various encoder backbones, demonstrating the flexibility of our pipeline structure. Moreover, since ResBlocks retains information at different scales and inherently captures spatial hierarchies in the visual input, it results in a lead in PSNR, SSIM, and CSIM compared to ViT. Considering the parameter and data efficiency, we utilize ResBlocks as the Visual Encoder backbone.

\section{Conclusion}
This paper addresses the challenge of generating talking faces through the FT2TF pipeline, which uses first-person text captions to replace traditional audio input. Our approach ensures the generation of faces with natural and detailed expressions. By incorporating two large language models, we extract emotional and linguistic features from the text. Moreover, the Multi-Scale Cross-Attention module effectively fuses visual and textual features, contributing to the synthesis of emotionally expressed talking faces. Extensive experiments on the LRS2 and LRS3 datasets highlight FT2TF's effectiveness in generating realistic, natural, and emotionally expressive talking faces from first-person text.

\renewcommand\thesection{\Alph{section}}
\setcounter{section}{0}
\renewcommand{\thetable}{S\arabic{table}}  
\renewcommand{\thefigure}{S\arabic{figure}}

\section{Implementation Details}
\label{sec:imple_details}

\subsection{Video Preprocessing}
For video data preprocessing, we employ the facial cropping method introduced in \cite{prajwal2020lip}. To further enhance computational efficiency, we standardize the input frame resolution to 96×96 pixels, balancing detail retention with processing speed.

\subsection{Audio Preprocessing}
In alignment with \cite{prajwal2020lip}, we generate Mel-spectrograms from audio samples recorded at 16kHz. This approach involves setting a window size of 800 samples and a hop size of 200, ensuring temporal resolution appropriate for capturing the nuances of lip movements.

\subsection{Visual Encoder}
The Visual Encoder plays a critical role in modeling spatial features from the talking face video. It consists of a deep convolutional architecture with 18 2D convolutional layers, each accompanied by Batch Normalization and ReLU activation functions. In addition, a subset of these modules incorporates residual blocks and skip connections, enhancing its ability to model complex visual features.

\subsection{Global Emotion Text Encoder}
The Global Emotion Text Encoder leverages a pretrained Emoberta \cite{kim2021emoberta} model to encode the overarching emotional tones within the caption text. We import \textit{SentenceTransformer} from the \textit{sentence\_transformers} library in Python and specifically use the \textit{tae898/emoberta-base} model. 

\subsection{Linguistic Text Encoder}
We adopt a pretrained GPT-Neo~\cite{gao2020pile} model as our Linguistic Text Encoder. The encoder is set up using the \textit{transformers} library in Python. We import  \textit{AutoTokenizer} and \textit{GPTNeoModel} from \textit{transformers} library. The tokenizer is initialized with \textit{AutoTokenizer.from\_pretrained("EleutherAI/gpt-neo-2.7B")}, and the model is initialized with \textit{GPTNeoModel.from\_pretrained("EleutherAI/gpt-neo-2.7B")}, respectively.

\subsection{Visual Decoder}
The Visual Decoder is composed of six sets of 2D transpose convolution blocks, with each block consisting of a 2D transpose convolution layer and two layers of 2D residual convolution blocks. This architectural design enables the transformation of text descriptions into a coherent and expressive sequence of video frames.

\subsection{Loss Weight Tuning}
\(\lambda_{1}\), \(\lambda_{2}\), and \(\lambda_{3}\) are the weight of three losses, fine-tuned for optimal performance. During the initial 300 epochs, we assign 0.7 to \(\lambda_{1}\), 0.09 to \(\lambda_{2}\), and 0.21 to \(\lambda_{3}\). Afterward, we adjust these weights, setting \(\lambda_{1}\) to 0.9, \(\lambda_{2}\) to 0.03, and \(\lambda_{3}\) to 0.07.

\section{Loss Structural Insights} 
We explore the structural aspects of the Face Synthesizer and the Discriminator, highlighting their respective contributions and characteristics.

The Face Synthesizer in FT2TF is inspired by the frozen-weight, pretrained Syncnet model \cite{prajwal2020lip}. It serves as a critical component in the lip synchronization process by aligning the synthesized talking face frames with the corresponding audio data. While not explicitly detailed through mathematical expressions, the Face Synthesizer integrates audio information to enhance the naturalness of lip movements in the generated talking face frames.

The Discriminator is composed of 14 layers of fully convolutional modules, without the use of Batch Normalization layers or skip connections. It operates as a binary classifier, providing binary predictions (\(O_{pred}\)) based on whether the input is generated or Ground Truth. These predictions are guided by the binary labels (\(Y_{disc}\)). This binary classification loss, as detailed in \(L_{disc}\), helps to ensure the quality and authenticity of the synthesized frames.

In summary, FT2TF's loss functions, coupled with the structural characteristics of the Face Synthesizer and the Discriminator, facilitate a comprehensive training process. These components work in harmony to improve the pixel-level fidelity, lip synchronization, and overall realism of the generated talking face videos, making FT2TF an effective solution for natural and expressive talking face synthesis.

\begin{figure*}[!ht]%[tbhp]
  \centering
  \includegraphics[width=140mm]{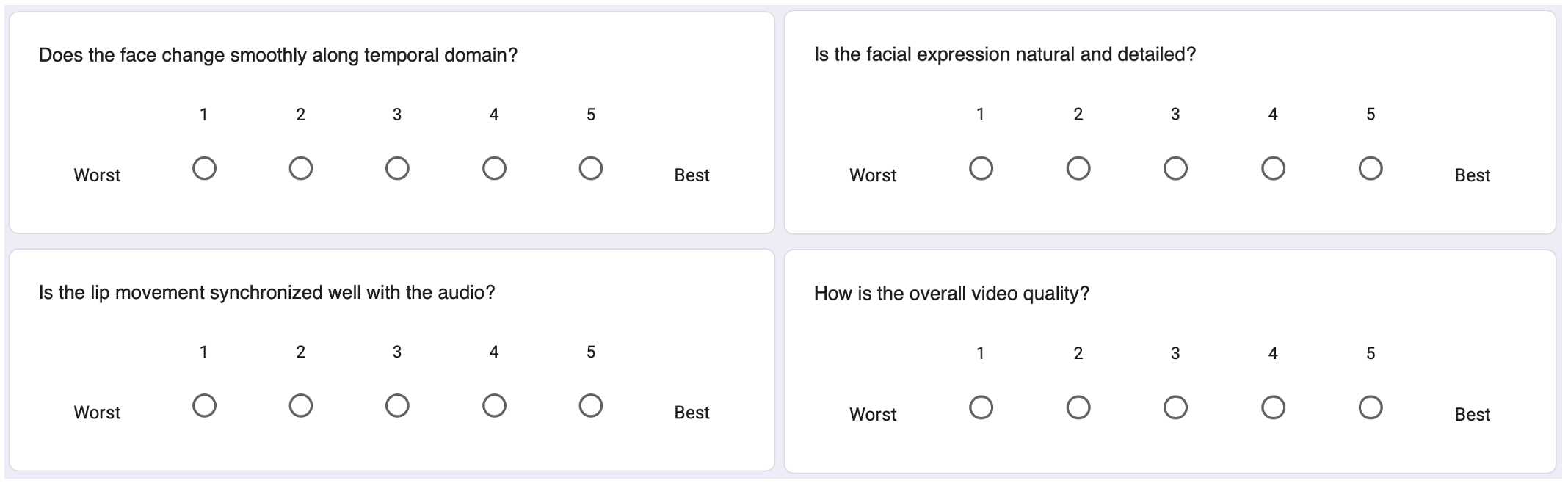}
  \caption{\textbf{User study questions.} 
We design four questions for each method in the user study to evaluate the quality of generated talking faces across multiple dimensions.
  }
  \label{fig:user_question}
\end{figure*}

\begin{figure*}[!ht]%[tbhp]
  \centering
  \begin{subfigure}{\linewidth}
    \includegraphics[width=\textwidth]{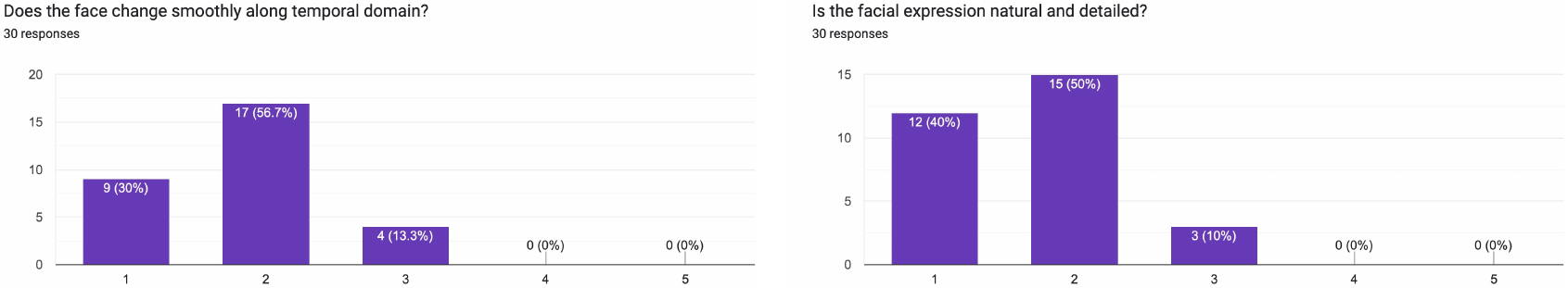}
    \caption{Users' responses of MakeItTalk \cite{zhou2020makelttalk}.}
    \label{fig:gl_att}
  \end{subfigure}
  \hfill
  \begin{subfigure}{\linewidth}
    \includegraphics[width=\textwidth]{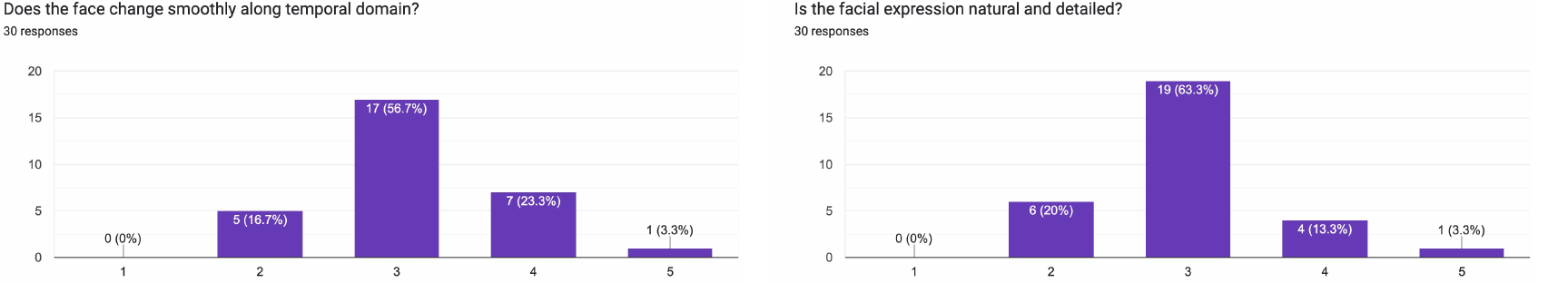}
      \caption{Users' responses of IP\_LAP \cite{zhong2023identity}.}
    \label{fig:lc_att}
  \end{subfigure}
    \begin{subfigure}{\linewidth}
    \includegraphics[width=\textwidth]{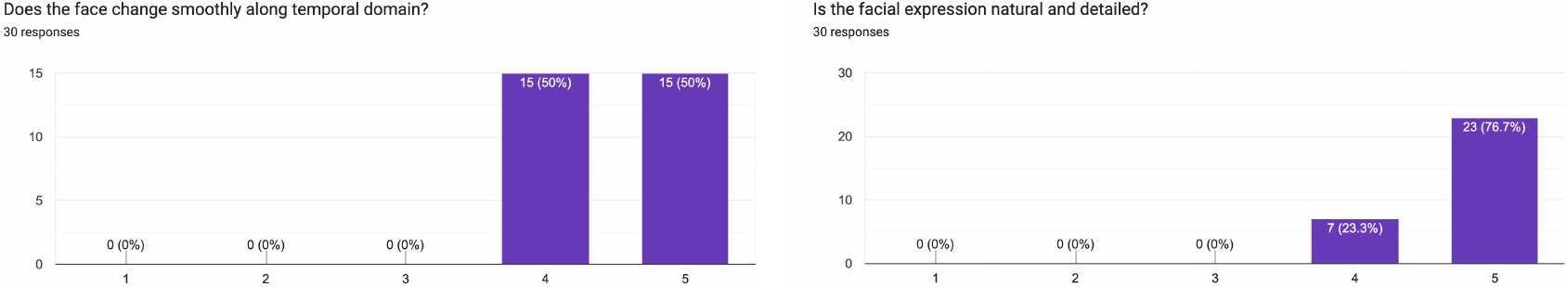}
      \caption{Users' responses of \textbf{FT2TF (ours)}.}
    \label{fig:lc_att}
  \end{subfigure}
  \caption{\textbf{Summary of user responses across methods.} The results indicate that user evaluations are notably more favorable for FT2TF compared to other methods.}
  \label{fig:response}
\end{figure*}

\section{User Study}

To evaluate the quality of FT2TF's generated talking faces in comparison to existing methods, we conduct a comprehensive user study assessing participant impressions across multiple dimensions.

\paragraph{Questionnaires.} The questionnaire used in the user study is shown in Figure \ref{fig:user_question}. Four questions are designed for each method, covering key aspects: temporal transition smoothness, lip synchronization accuracy, facial detail naturalness, and overall video quality.

\paragraph{User Responses.} 
Figure \ref{fig:response} provides a summary of user responses. The results indicate that participants responded significantly more positively to FT2TF compared to other methods.

{\small
\bibliographystyle{ieee_fullname}
\bibliography{egbib}
}

\end{document}